# Environmental large language model Evaluation (ELLE) dataset: A Benchmark for Evaluating Generative AI applications in Eco-environment Domain

Jing Guo [1], Nan Li [2], Ming Xu [2*]

[1] Beijing Information Science & Technology University

[2] Tsinghua University

*xu-ming@tsinghua.edu.cn

## Abstract

Generative artificial intelligence (AI) demonstrates significant potential in the ecological and environmental sectors, offering applications such as environmental monitoring, data analysis, educational tools, and policy support. However, the effectiveness of these applications remains contentious due to the absence of a unified and reliable evaluation framework to assess the professionalism and applicability of generative AI within this specialized domain. To address this gap, we present the Environmental large language model Evaluation (ELLE) question-answer (QA) dataset, i.e., the ELLE-QA Benchmark. It is the first dedicated dataset designed to evaluate large language models (LLMs) and their derivatives, including model fine-tuning and retrieval-augmented generation, specifically in the ecological and environmental sciences. Our research methodology involved the creation of 1,130 QA pairs sourced from expert questionnaire and open-source authoritative materials, covering 16 distinct environmental subjects to ensure comprehensive disciplinary coverage. These QA pairs were systematically categorized by content domain, difficulty level, and question type, forming the foundation of ELLE framework. The ELLE-QA Benchmark establishes a comprehensive and trustworthy evaluation system, advancing the standardization of AI assessments in the ecological and environmental fields. By providing the first dedicated dataset for testing AI models' proficiency in environmental and ecological knowledge, our benchmark supports consistent and objective comparisons, thereby promoting the robust development and application of AI technologies. This standardized evaluation framework is poised to drive advancements in ecological and environmental AI research, fostering sustainable and impactful outcomes in the field. The full ELLE-QA

Benchmark dataset and code are available at https://elle.ceeai.net/ and https://github.com/CEEAI/elle.

## 1. Introduction

Generative artificial intelligence (GenAI) technologies, exemplified by LLMs such as ChatGPT (OpenAI, 2025) and Claude (Anthropic, 2025), have achieved notable progress in natural language processing, enabling them to produce coherent, contextually relevant, and diverse textual outputs. This advancement has unlocked a wide range of applications across various domains, including the ecological and environmental fields, where such technologies can play a transformative role. For instance, AI models have been employed to analyze camera trap footage, drone imagery, and GPS data for wildlife tracking, habitat assessment, and biodiversity analysis. This approach improves anti-poaching efforts and the protection of diverse species (Chisom et al., 2024). Additionally, in marine biodiversity monitoring, generative AI enables scalable and adaptable species identification by leveraging open-domain learning frameworks, such as pretrained vision-language models with retrieval-augmented generation (RAG). This approach effectively analyzes images and videos from dynamic and diverse ocean environments, identifying rare or unseen species without domain-specific training (Dyanatkar et al., 2024). LLM agents also show promise in participatory urban planning, addressing traditional challenges of time and resource constraints. A recent study demonstrated this by developing a specialized LLM-based framework that could handle land-use planning tasks through role-playing, collaborative generation, and feedback iteration. When tested across diverse urban communities, the LLM-based system not only surpassed human experts in stakeholder satisfaction and inclusiveness metrics but also achieved comparable performance to state-of-the-art reinforcement learning methods in service delivery and ecological considerations (Zhou et al., 2024).

Despite the evident potential, the application effectiveness of generative AI in the ecological environment domain remains a subject of ongoing debate. One of the primary obstacles hindering the widespread adoption and optimization of these technologies is the absence of a standardized and reliable evaluation framework. Current assessments often lack the specificity required to measure the professionalism and applicability of generative AI within specialized fields such as ecology and environmental science. This gap underscores the need for standardized benchmarks that can accurately reflect the performance of AI models in handling domain-specific tasks and providing actionable insights.

In response to this critical need, we present the ELLE-QA Benchmark, a pioneering dataset meticulously designed to evaluate the capabilities of large language models and their applications within the ecological and environmental domain. The ELLE-QA Benchmark aims to bridge the existing evaluation gap by offering a comprehensive and reliable assessment system tailored to the unique demands of ecological and environmental sciences. By establishing a benchmark QA dataset, we provide a structured means to measure and compare the performance of generative AI technologies across key evaluation dimensions: professionalism, clarity, and feasibility.

The establishment of the ELLE-QA Benchmark represents a significant advancement in the standardization of AI evaluation within the ecological and environmental domains. By providing a unified comparison benchmark, the dataset facilitates consistent and objective assessments, enabling researchers and practitioners to benchmark AI performance accurately. This, in turn, supports the development of more robust and effective AI solutions tailored to the intricate challenges of ecological and environmental management.

## 2. Related works

In recent years, LLMs have been deployed in many fields, and important technical innovations such as Retrieval-Augmented Generation (RAG) have enabled LLMs to more effectively integrate external knowledge, while fine-tuning on task-specific datasets has allowed them to adapt to specialized tasks. As a result, new frontiers in performance have been reached, with LLMs now capable of managing complex reasoning, providing context-rich responses, and supporting domain-specific inquiries. Given these rapid developments, the evaluation of LLMs has likewise become more sophisticated, as researchers and industry practitioners seek comprehensive benchmarks that can reliably measure model performance across both general and specialized domains.

**General LLM Evaluation**

One prominent line of work involves evaluating general-purpose models, or "foundation models," across diverse tasks and languages. In the Chinese NLP community, SuperCLUE serves as a representative example of this approach (Xu et al., 2023). As an extension of the CLUE benchmark for Chinese language understanding, SuperCLUE tests a wide array of capabilities—basic skills, professional expertise, and unique Chinese linguistic characteristics—across various models, including internationally recognized ones. The goal is to provide insights into questions such as "How do these Chinese LLMs compare to leading international models?" and "How do they measure up against human performance?"

Similarly, another LLM evaluation dataset, the JioNLP, developed by a lab in China, targets general LLM effectiveness by focusing on real-world tasks (JioNLP, 2025). It includes both objective and subjective components, with multiple-choice questions derived from professional exams in China (covering about 32% of the dataset) to test knowledge coverage, and open-ended questions assessing common practical functions of LLMs.

Other examples of general-purpose model evaluation include C-Eval, a multi-disciplinary Chinese benchmark suite for large language models, and the Hugging Face model evaluation frameworks, which offer tools to benchmark a wide range of transformers-based models across standard tasks (Huang et al., 2024). These efforts collectively highlight the importance of comprehensive and fair evaluation methodologies for understanding how models perform across different tasks, languages, and formats.

**Vertical Domain Model Evaluation**

With the growing prominence of specialized LLM applications, researchers have also sought to design domain-specific evaluations. For example, in user-centered chat scenarios, one evaluation framework divides the test into objective and subjective sections. The objective section comprises around 50 common queries that frequently arise in voice-based chat systems, and outputs from models are rated on quality and relevance by GPT-4.0 or similar services. This combination of curated prompts and automated scoring metrics provides a practical gauge for model performance in real-world, domain-specific settings.

In the financial domain, an automated and comprehensive benchmark called OmniEval has been introduced to assess RAG systems (Wang et al., 2024). OmniEval examines various dimensions of information retrieval and generation relevant to finance, thereby offering a specialized lens on domain-specific competencies and challenges. Likewise, in the biomedical domain, BioMistral demonstrates the efficacy of adopting a foundation model (Mistral), further pretraining it on PubMed Central data, and evaluating it on ten established English medical QA tasks (Labrak et al., 2024). The results show that BioMistral outperforms existing open-source medical models and remains competitive against proprietary ones. The research team even extended the benchmark to seven additional languages, creating the first large-scale multilingual LLM evaluation in the medical field.

Despite these advances, there remains a significant gap in evaluating the performance of RAG models within the ecological environment domain. No unified or reliable framework currently exists to measure the professional applicability and rigor of such models in this area. This gap underscores the necessity of constructing a high-quality, comprehensive benchmark to assess RAG models' capabilities in handling complex, domain-specific ecological issues. Our study aims to address this challenge by establishing and validating a specialized evaluation framework tailored to the ecological environment domain.

## 3. Data Collection

### 3.1 Framework and principles of ELLE-QA Benchmark

The development of the ELLE-QA Benchmark involved the creation of a robust methodological framework that guides models in generating domain-specific professional questions. This framework ensures that the questions are not only relevant and challenging but also reflective of the multifaceted nature of ecological and environmental issues. The benchmark encompasses a wide range of topics, including environmental geology, chemistry, ecology, toxicology, and management, among others, thereby covering the breadth and depth necessary for a thorough evaluation.

At its core, the ELLE-QA Benchmark incorporates three critical evaluation dimensions to assess generative AI performance comprehensively. Professionalism evaluates the model's ability to generate accurate and domain-relevant content, ensuring that the information provided aligns with established scientific knowledge and practices. Clarity assesses the model's proficiency in articulating responses that are clear, concise, and easily understandable, which is essential for effective communication in both academic and practical applications. Feasibility examines the practicality and applicability of the model's outputs, determining whether the generated solutions and analyses are viable within real-world ecological and environmental contexts.

Stringent rules and format specifications were established for the collection of each QA pair to ensure the benchmark QA set is both comprehensive and robust. Every QA pair not only includes the question and its corresponding answer but also provides essential metadata detailing the content scope, difficulty level, and question type. This structured approach facilitates a systematic evaluation of large language models within the ecological environment domain. The benchmark QA set aims to encompass the full spectrum of the ecological environment field, ensuring extensive coverage across various specialized sub-disciplines. The difficulty of each QA pair is categorized into three distinct levels—Simple, Medium, and Hard—based on well-defined scientific and practical principles. To capture the diverse cognitive demands placed on LLMs, QA pairs are categorized into three primary types: knowledge, calculation, and reasoning. The specific definition and summary of each category are shown in Table 1.

Table 1 Summary table of QA pair specifications

| *Category* | *Description* |
| --- | --- |
| *Content domains* | Environmental Geology, Environmental Chemistry, Environmental Ecology, Environmental Mathematics, Environmental Toxicology, Environmental Physics, Water |
| | Environment, Atmospheric Environment, Soil Environment, Biological Environment, Environmental Engineering, Environmental Control, Environmental Monitoring, Environmental Law, Environmental Economics, Environmental Management, Environmental Ethics, etc. |
| *Difficulty levels* | **Simple:** Basic concepts and terminology with direct answers.<br>**Medium:** Integration of multiple concepts, basic data interpretation, or case studies.<br>**Hard:** Complex issues requiring advanced analysis, interdisciplinary knowledge, and comprehensive solutions. |
| *Question types* | **Knowledge:** Fundamental and advanced knowledge, terminology, historical and recent developments.<br>**Calculation:** Mathematical operations, data analysis, quantitative problem-solving.<br>**Reasoning:** Analytical and systemic thinking, conditional reasoning, causal and analogical reasoning, hypothesis testing. |

### 3.2 Questionnaire-based QA pairs collection

To construct a comprehensive and representative benchmark QA set for evaluating LLMs in the ecological environment domain, a structured questionnaire approach was employed. A diverse and extensive group of experts was assembled to ensure the QA pairs encompass the multifaceted aspects of the ecological environment. These experts hailed from various relevant fields, including but not limited to ecology, environmental science, data science, artificial intelligence, and statistics. The selection criteria prioritized individuals with substantial experience and recognized expertise in their respective domains to guarantee the quality and reliability of the generated content.

The questionnaire was meticulously designed to solicit medium to hard level questions pertinent to the ecological environment. Multiple distribution channels were utilized to maximize reach and participation, including professional networks, social platforms, and specialized online platforms. Participants were invited to contribute by submitting both challenging questions and their corresponding answers, thereby ensuring that the collected data would reflect a broad spectrum of expertise and inquiry levels. To facilitate ease of response and encourage comprehensive participation, the questionnaire was deployed through an online survey tool. Regular follow-ups and reminders were sent to maintain engagement and ensure sufficient quantity of QA pairs were achieved. The initial phase aimed to gather a substantial dataset, and contingent upon the progress and quality of submissions, a second round of questionnaires was planned to further enrich the dataset if necessary.

### 3.3 Manual collection of open-source QA pairs

Complementing the questionnaire-based approach, a portion of the benchmark QA set was manually curated from various open-source materials. This process focused on gathering medium-difficulty professional questions and their corresponding answers within the ecological environment domain, ensuring a comprehensive and diverse dataset. The primary sources for this manual collection included both English and Chinese environmental science textbooks, which provided foundational and advanced topics across multiple sub-disciplines. Additionally, past examination question sets from environmental science courses and certification exams were reviewed to extract standardized questions that effectively assess different levels of understanding and application. Professional consultations constituted another vital source, where queries submitted to environmental departments and the expert responses received were incorporated to reflect real-world problem-solving scenarios.

The manual collection process was systematically structured to ensure the relevance, accuracy, and diversity of the QA pairs. Each data source was thoroughly reviewed to identify questions that align with medium difficulty levels, emphasizing relevance to key ecological concepts, applicability to real-world scenarios, and the capacity to evaluate critical thinking and problem-solving skills. A bilingual inclusion strategy was employed, incorporating both English and Chinese materials to capture a broader spectrum of perspectives and terminologies, thereby enhancing the dataset's applicability in diverse linguistic contexts.

### 3.4 Cross-screening and validation of QA pairs

Following the comprehensive collection of QA pairs through questionnaires and manual sourcing, a rigorous cross-screening and validation process was undertaken to ensure the quality, relevance, and reliability of the benchmark dataset. As mentioned above, all collected QA pairs were meticulously categorized based on three primary dimensions: professional domain, difficulty level, and question type. Based on this classification, an initial screening was conducted to eliminate any QA pairs that did not meet the predefined criteria for relevance, clarity, and appropriateness. This preliminary filtering aimed to streamline the dataset by removing redundant, overly simplistic, or off-topic pairs, thereby enhancing the overall quality and focus of the collection, and ensuring that each remaining QA pair contributed uniquely to the dataset's overall comprehensiveness.

To further ensure the dataset's robustness, a specialized expert panel was convened to undertake a three-round cross-review process. In each round, subsets of the QA pairs were independently evaluated by different groups of experts to assess their relevance to the ecological environment domain, appropriate difficulty level, and correct classification of question types and professional fields. During the cross-review process, any QA pairs that elicited uncertainty or disagreement among reviewers were flagged for further examination. These contentious pairs were subsequently deliberated in dedicated expert review meetings, where panel members engaged in in-depth discussions to resolve disputes and reach a consensus on their suitability for inclusion in the final benchmark set. This collaborative decision-making process ensured that only those QA pairs meeting the highest standards of scientific accuracy and relevance were retained.

## 4. Generated dataset: characteristics and composition

A total of 1130 QA pairs were generated to evaluate GenAI applications in the ecological and environmental sciences domain, and a total of 16 subjects were covered. The disciplinary coverage reveals nuanced distribution, with approximately 23.37% of the QA pairs categorized under environmental ecology, followed by environmental engineering with a proportion of 18.67%. The statistical distribution results across various environmental subject areas are illustrated in Fig. 1.

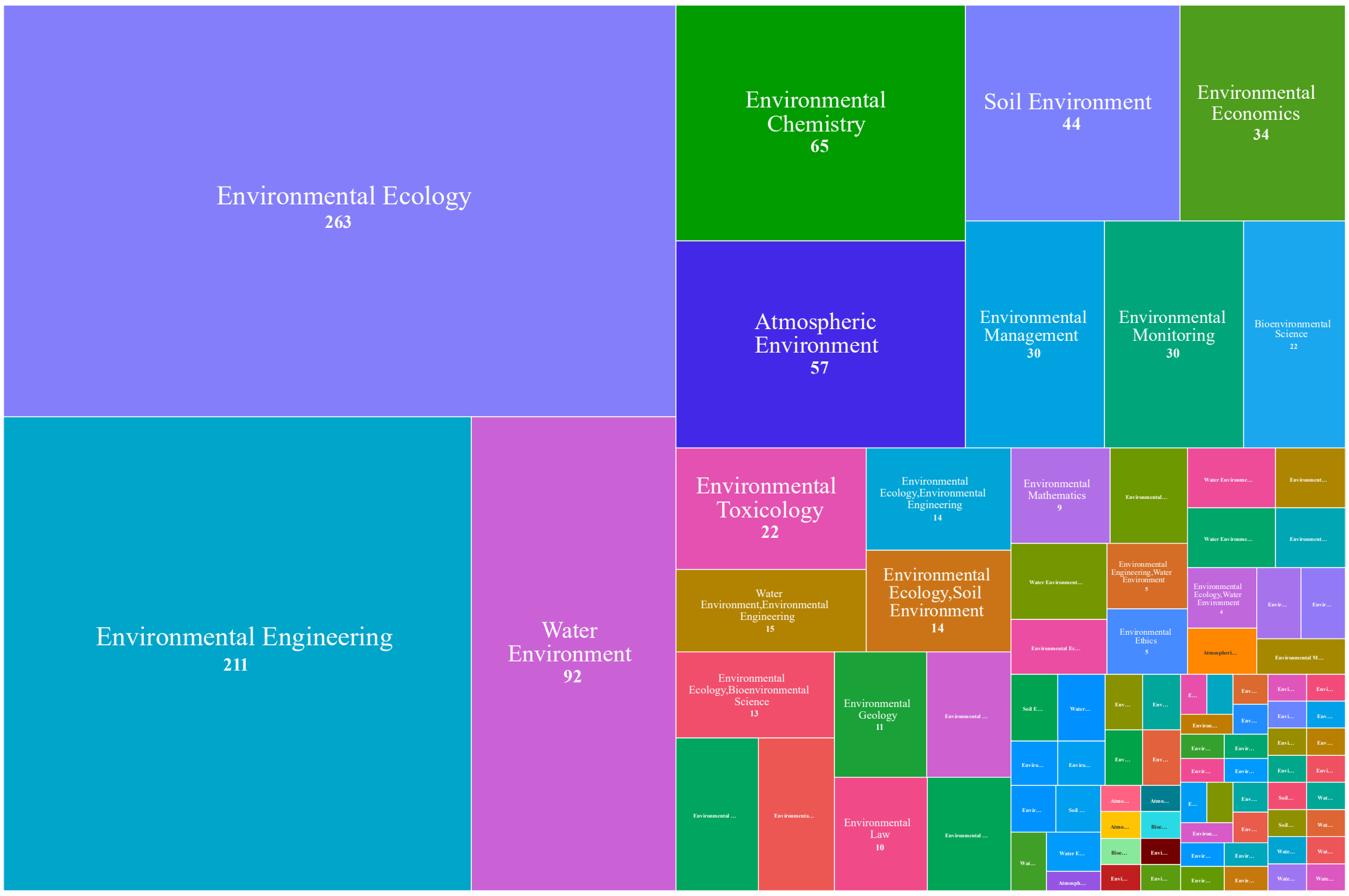

Fig. 1 The statistical distribution results across various environmental subjects

Nevertheless, the dataset's typology demonstrates diversity in types and complexity levels. Knowledge questions accounted for the majority, with 565 questions (50%) as the base category, and an additional 101 questions combining with reasoning, 25 with calculation, and 7 questions incorporating all three types. Reasoning questions represented the second largest category with 325 base questions, accounting for 28.76%, along with 21 questions merged with calculation. In contrast, calculation questions the smallest portion, with 86 base questions, comprising 7.61% of the generated QA pairs. In terms of difficulty levels, 43.10% of QA pairs were categorized as medium-level, followed by hard-level with 431 questions (38.14%). Easy questions were minimal, contributing only 18.76%. The distribution results of difficulty levels and types in ELLE-QA Benchmark are shown in Fig. 2.

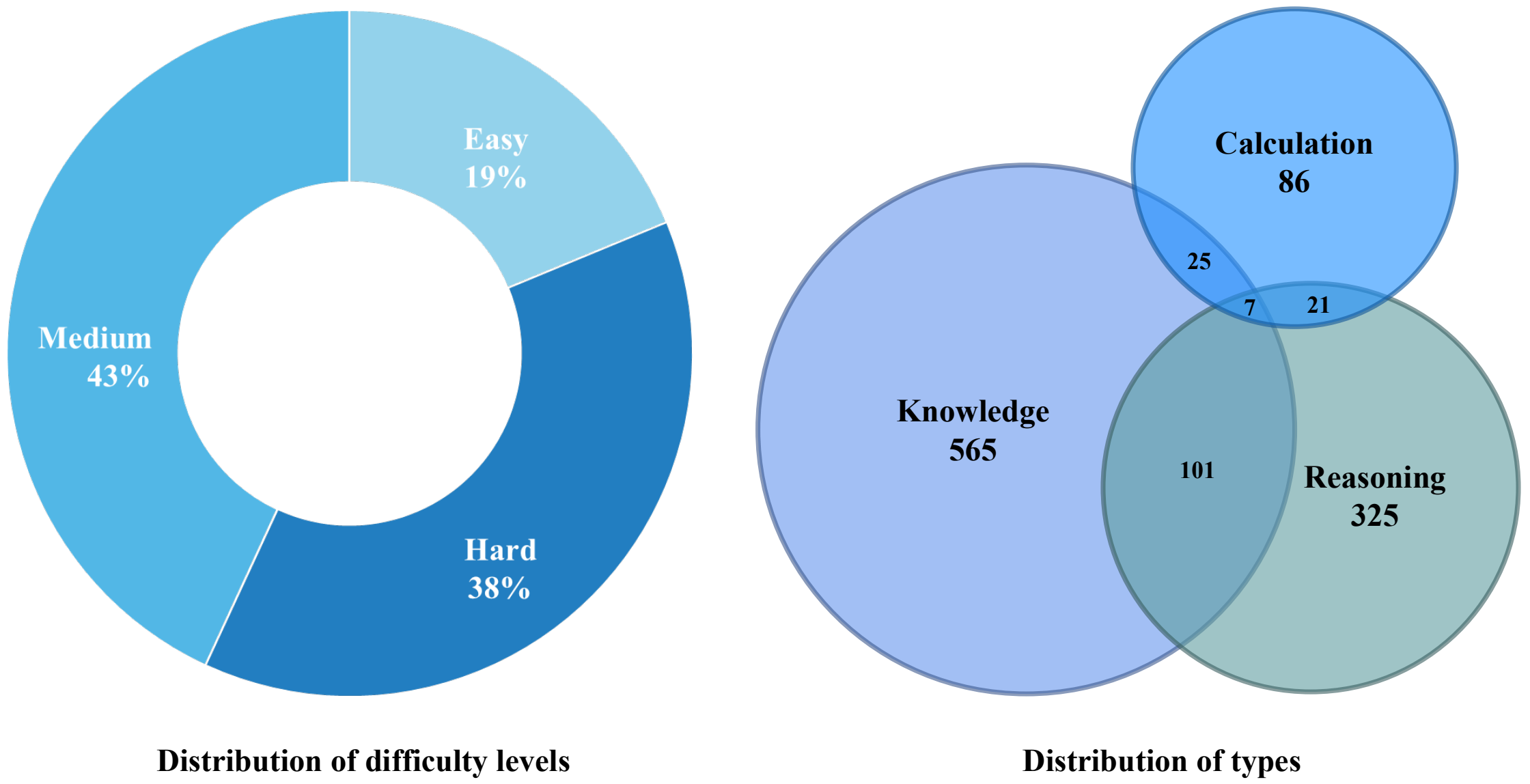

Fig. 2 The distribution results of difficulty levels and types in ELLE-QA Benchmark

A detailed breakdown of question difficulty and type distribution reveals distinct patterns. Reasoning questions exhibited the highest concentration of hard-level items, with 186 out of 325 questions falling into this category. Questions of knowledge were more evenly distributed, with 156 easy-level, 275 medium-level and 134 hard-level questions, indicating diverse complexity within this type. Conversely, calculation questions, though fewer, skewed toward higher difficulty levels, with 39 hard, 27 medium, and 20 easy questions.

This nuanced distribution reveals a deliberate and strategic approach to dataset construction, where the concentration of challenging reasoning, calculation, and knowledge questions at high difficulty levels creates a rigorous testing ground that demands sophisticated analytical capabilities, deep domain expertise, and complex problem-solving skills from LLMs in the environmental ecology domain. This emphasis on critical thinking and problem-solving skills aligns with the prompt design to generate challenging questions for effectively assessing advanced model performance.

## 5. Evaluation protocol and application of ELLE-QA Benchmark

The ELLE-QA Benchmark is designed to facilitate rigorous and standardized evaluations of LLMs within the ecological and environmental domains. To maintain the integrity and objectivity of the evaluation process, the benchmark exclusively publishes the questions along with their classifications, which include the specific environmental discipline, difficulty level, and question type. The answers to these questions are withheld to prevent any bias during the assessment phase.

When utilizing the benchmark to evaluate different LLMs, researchers and practitioners must adhere to the established evaluation criteria depicted in Table 2. The evaluation and scoring process adopts a hybrid approach, combining AI with human expert assessments. Each model is assessed across the three primary dimensions of professionalism, clarity, and feasibility. For each dimension, models receive individual scores based on their performance in generating accurate, clear, and practical responses to the provided questions. These scores are then aggregated to produce a comprehensive performance metric for each model, enabling a holistic comparison of their capabilities.

Table 2 The evaluation criteria of ELLE

| *Evaluation Dimensions* | *Knowledge* | *Reasoning* | *Calculation* |
|---|---|---|---|
| *Accuracy* | Aligns with authoritative knowledge or standard conclusions, avoiding factual errors. | Conclusions align with logical rules, avoiding fallacies. | Outcomes are accurate, consistent with standard answers, and account for boundary conditions. |
| *Logical consistency* | Derivations are rigorous, with clear logic, avoiding skipped steps or ambiguous reasoning. | Step-by-step progression, fully utilizing known conditions with reasonable explanations. | Steps are complete and clear, using efficient methods with well-defined assumptions and constraints. |
| *Normative expression* | Terminology is precise, language is concise and | Coherent and understandable | Mathematical expressions are |
| | clear, adhering to professional formats. | presentation, with clear logic and strong readability. | standardized, variables are well-defined, and answers are clearly summarized. |

To ensure transparency and continuous improvement, the standard answers to the benchmark questions are released only after the evaluation results of the models have been published. This approach allows for an unbiased assessment of the models' performance while providing the necessary reference points for understanding and interpreting the scores. Additionally, the ELLE-QA Benchmark employs a leaderboard system, where evaluation results of various models are periodically updated in different "seasons." This dynamic updating mechanism ensures that the benchmark remains current with the latest advancements in LLM technologies and provides an ongoing platform for tracking and comparing model performance over time.

By following this structured evaluation protocol, the ELLE-QA Benchmark not only standardizes the assessment of genAI models in the ecological and environmental fields but also fosters a competitive and progressive environment for AI development. The continual updates and transparent sharing of standard answers further enhance the benchmark’s utility, making it an essential tool for advancing the application and effectiveness of AI technologies in addressing complex ecological challenges.

## 6. Conclusion

We introduced the ELLE-QA Benchmark, a novel and comprehensive dataset specifically designed to evaluate the performance and applicability of large language models within the ecological and environmental domains. Recognizing the burgeoning potential of generative artificial intelligence technologies, in addressing complex ecological challenges, our work addresses the critical need for a standardized and reliable evaluation framework. The absence of such benchmarks has previously hindered the objective assessment and optimization of AI models for specialized fields, limiting their effective deployment and integration into ecological and environmental applications.

The development of the ELLE-QA Benchmark was underpinned by a meticulous data collection process, which encompassed both questionnaire-based and manual sourcing methods, to ensure that the collected QA pairs were not only extensive in quantity but also rich in depth and breadth. The inclusion of questions from a wide array of environmental fields, ranging from environmental geology and chemistry to environmental ethics and management, guarantees that the benchmark captures the multifaceted nature of ecological issues. Additionally, the rigorous cross-screening and validation process, involving multiple rounds of expert reviews and consensus-building discussions, further reinforced the dataset’s scientific integrity and reliability.

Our structured approach to categorizing QA pairs based on content domains, difficulty levels, and question types facilitates a nuanced evaluation of language models. By defining clear and practical guidelines for difficulty classification and question typology, we enable a detailed analysis of AI performance across different cognitive demands and specialized areas. This ensures that the ELLE-QA Benchmark not only assesses the factual accuracy and clarity of generated responses but also evaluates the models' ability to perform complex reasoning and problem-solving tasks relevant to the ecological environment sector.

Preliminary assessments using the ELLE-QA Benchmark have yielded insightful results, highlighting both the strengths and areas for improvement in current generative AI models. These initial findings provide a foundation for ongoing analysis and refinement, offering valuable feedback to enhance the models’ capabilities and ensure their alignment with the specialized needs of the ecological environment sector.

**Acknowledgements**

We thank the following contributors for their efforts in developing the ELLE-QA Benchmark dataset, listed in alphabetical order by their first names.

| Contributors | Affiliated institution |
|---|---|
| Ayazhan Nurpeiis | Tsinghua University |
| Bin Zhu | Huawei Technologies Co., Ltd. |
| Changqing Xu | Beijing Institute of Technology |
| Chao Zhang | Tsinghua University |
| Chen Qian | University of Science and Technology of China |
| Cheng Gong | Shanghai NIO Automobile Co., Ltd. |
| Chenling Fu | Tsinghua University |
| Chenyang Shuai | Chongqing University |
| Chuanzhong Chen | China National Environmental Monitoring Centre |
| Chuke Chen | Tsinghua University |
| Chunyan Wang | Tsinghua University |
| Dahai Meng | Shanghai NIO Automobile Co., Ltd. |
| Di Wu | China Construction Intelligent Technology Co., Ltd. |
| Dongxu Zhang | CRRC Zhuzhou Institute Co., Ltd. |
| Fan Lv | Tongji University |
| Fei Jiang | Nanjing University |
| Gao Chen | Jiangsu Provincial Eco-Environmental Monitoring and Supervision Company |
| Gongliang Zhang | Beijing Capital Eco-Environmental Protection Group Co., Ltd. |
| Guodong Xu | CSD Water Service Co., Ltd. |
| Guoguo Liu | Schneider Electric (China) Co., Ltd. |
| Guoqiang Qian | Beijing Zhongchuang Carbon Investment Technology Co., Ltd. |
| Hang Yang | Tsinghua University |
| Hanyuan Wang | Climind Company |
| Haochun Yan | China National Inspection and Testing Holdings Co., Ltd. |
| Heng Liang | Harbin Institute of Technology |
| Hongbin Liu | Nanjing Forestry University |
| Hongcheng Wang | Harbin Institute of Technology |
| Honggui Han | Beijing University of Technology |
| Hongliang Zhang | Fudan University |
| Huacheng Wu | North China (Jibei) Electric Power Research Institute |
| Hui Chen | China Southern Power Grid Carbon Asset Management Co., Ltd. |
| Huimin Chang | Tsinghua University |
| Jia He | Shanghai Zhengsheng Cloud Computing Co., Ltd. |
| Jianchuan Qi | Tsinghua University |
| Jiang Bian | Microsoft Research Asia |
| Jianguo Tian | School of Business, University of Jinan |
| Jianhua Mao | Beijing Enterprises Water Group Limited |
| Jianwei Du | South China Institute of Environmental Sciences, Ministry of Ecology and Environment |
| Jianxi Luo | City University of Hong Kong |
| Jiaqi Lu | Shanghai University of Engineering Science |
| Jiayi Yuan | Tsinghua University |
| Jinfeng Wang | Nanjing University |
| Jinliang Xie | Tsinghua University |
| Jinyi Tian | East China University of Science and Technology |
| Jiping Jiang | Southern University of Science and Technology |
| Jiwei Wu | Sichuan University |
| Juan Ma | Research Center for Eco-Environmental Sciences, Chinese Academy of Sciences |
| Jue Liu | Peking University |
| Jun Lv | iFLYTEK Co., Ltd. |
| Jun Wan | Chinese Academy of Environmental Planning, Ministry of Ecology and Environment |
| Junfeng Wang | Nanjing University of Information Science and Technology |
| Lei Shi | Nanchang University |
| Liang Zhang | Beijing University of Technology |
| Liangzhi Li | Climind Company |
| Lijin Zhong | Beijing Huan Ding Environmental Big Data Research Institute |
| Lin Wang | Institute of Urban Environment, Chinese Academy of Sciences |
| Lin Qiu | Envision Digital Co., Ltd. |
| Miaomiao Liu | Nanjing University |
| Ming Xue | Safety and Environmental Protection Technology Research Institute Co., Ltd., China National Petroleum Corporation |
| Mingzhi Huang | South China Normal University |
| Na Zhang | Minviro Shanghai |
| Peng Li | State Grid Jibei Electric Power Research Institute |
| Pengbo Fu | East China University of Science and Technology |
| Qi Chen | School of Environmental Sciences and Engineering, Peking University |
| Qiang Yang | Research Institute of East China University of Science and Technology |
| Qili Dai | Nankai University |
| Qiuwan Wang | Beijing Baidu Netcom Science and Technology Co., Ltd. |
| Ran Cai | Beijing Capital Eco-Environmental Protection Group |
| Rentao Ouyang | Tsinghua University |
| Ruijun Zhang | Hebei University of Technology |
| Ruirui Zhang | Tsinghua University |
| Runlong Hao | North China Electric Power University |
| Ruoxi Xiong | Tsinghua University |
| Ruru Han | Tsinghua University |
| Shangheng Yao | China Southern Power Grid Energy Development Research |

| | Institute |
|---|---|
| Shen Qu | Beijing Institute of Technology |
| Shenggui Ma | Tianfu Yongxing Laboratory |
| Shifa Zhong | East China Normal University |
| Shijie Cao | Southeast University |
| Shixin He | Harbin Institute of Technology |
| Shunyao Wang | Shanghai University |
| Shuwen Wang | CITIC Investment Holdings Co., Ltd. |
| Si Zhang | Tsinghua University |
| Sijie Lin | Tongji University |
| Simeng Chen | Tsinghua University |
| Sitong Liu | Peking University |
| Siyu Chen | Lanzhou University |
| Tao Sun | Tianjin Academy of Eco-Environmental Sciences |
| Tianhong Li | School of Environmental Sciences and Engineering, Peking University |
| Tong Sha | Shaanxi University of Science and Technology |
| Wanglai Cen | Sichuan University |
| Wei Zhang | Chinese Academy of Environmental Planning, Ministry of Ecology and Environment |
| Wei Li | Information Center, Ministry of Ecology and Environment |
| Wei Wang | Chinese Research Academy of Environmental Sciences |
| Wei Liu | Alibaba Group |
| Wei Wei | Beijing University of Technology |
| Weijun Zhang | Research Center for Eco-Environmental Sciences, Chinese Academy of Sciences |
| Weijun Li | Zhejiang University |
| Weiqi He | Research Institute for Environmental Innovation (Suzhou) Tsinghua |
| Wen Fang | Nanjing University |
| Wenhao Chen | China Mobile IoT Company Limited |
| Wenjie Shi | Tsinghua University |
| Xi Tian | Nanchang University |
| Xia Meng | Fudan University |
| Xiangzhong Guo | Institute of Urban Environment, Chinese Academy of Sciences |
| Xiaobin Tang | Harbin Institute of Technology |
| Xiaohui Lu | Tsinghua University |
| Xiaonan Wang | Tsinghua University |
| Xiaoxin Cao | China Water Environment Group |
| Xin Wang | Nankai University |
| Xin Dong | Tsinghua University |
| Xuedian Gu | China Resources Environmental Services Co., Ltd. |
| Xuehua Li | Dalian University of Technology |
| Xuelin Zhang | Sun Yat-sen University |
| Yan Wu | Jiangsu Skytech Industrial Internet Co., Ltd. |
| Yang Ou | Peking University |
| Yangyang Guo | Institute of Process Engineering, Chinese Academy of Sciences |
| Yesong Gao | China Construction Eco-Environment Group Co., Ltd. |
| Yihang Zhou | Tsinghua University |
| Yixin Zhu | Peking University |
| Yu Bai | Beijing Drainage Group Co., Ltd. |
| Yuanyi Huang | Cloud&Information (Guangdong) Eco-Environment Science and Technology Co., Ltd. |
| Yuehong Zhao | Institute of Process Engineering, Chinese Academy of Sciences |
| Yun Zhu | South China University of Technology |
| Yunduo Lu | Tsinghua University |
| Yuqiang Zhang | Institute of Environmental Research, Shandong University |
| Yutao Wang | Fudan University |
| Yuzhen Feng | Tsinghua University |
| Zhanjun Cheng | Tianjin University |
| Zhaoxin Dong | South China University of Technology |
| Zhe Wang | The Hong Kong University of Science and Technology |
| Zhe Jiang | Tianjin University |
| Zhen Cheng | Shanghai Jiao Tong University |
| Zheng Wang | Alibaba Group |
| Zhihua Li | Xi'an University of Architecture and Technology |
| Zhijun Gui | Shanghai E-Carbon Digital Technology Co., Ltd. |
| Zhongming Lu | The Hong Kong University of Science and Technology |
| Zimeng Cai | Tsinghua University |
| Ziqi Wang | Tsinghua University |
| Zongguo Wen | Tsinghua University |